\documentclass[final,12pt]{clear2023} % Include author names

\title[]{Pearl Causal Hierarchy on Image Data: Intricacies \& Challenges}
\usepackage{times}
\usepackage{wrapfig}
\usepackage{lipsum}  

\DeclareMathOperator{\doop}{\textit{do}}
\DeclareMathOperator{\pa}{pa}
\DeclareMathOperator{\Pa}{Pa}

\newcommand{\inner}[1]{\langle #1 \rangle}

\author{\vspace{.5cm}
\textbf{Matej Zečević}\textsuperscript{\rm 1,*}
\quad \vspace{-.5cm}
\textbf{Moritz Willig}\textsuperscript{\rm 1,*}
\quad \textbf{Devendra Singh Dhami}\textsuperscript{\rm 1,3} \quad \textbf{Kristian Kersting}\textsuperscript{\rm 1-4}\\
{\normalfont \textsuperscript{\rm 1}Computer Science Department, TU Darmstadt, \textsuperscript{\rm 2}Centre for Cognitive Science, TU Darmstadt, \textsuperscript{\rm 3}Hessian Center for AI (hessian.AI),
\textsuperscript{\rm 4}DFKI}
\vspace{-1cm}
}

\begin{document}

\maketitle

\begin{abstract}%
  Many researchers have voiced their support towards Pearl's counterfactual theory of causation as a stepping stone for AI/ML research's ultimate goal of intelligent systems. As in any other growing subfield, patience seems to be a virtue since significant progress on integrating notions from both fields takes time, yet, major challenges such as the lack of ground truth benchmarks or a unified perspective on classical problems such as computer vision seem to hinder the momentum of the research movement. This present work exemplifies how the Pearl Causal Hierarchy (PCH) can be understood on image data by providing insights on several intricacies but also challenges that naturally arise when applying key concepts from Pearlian causality to the study of image data.
\end{abstract}


\section{AI/ML Research Hopeful of the Promise given by Causality}
Speaking of causality, the Pearlian counterfactual theory of causation \citep{pearl2009causality} has recently found prominent support in the AI/ML community \citep{scholkopf2022causality, peters2017elements, geffner2022probabilistic}. An increasing presence of publications at major conferences/journals concerned with the integration of causality with AI/ML (including \citep{janzing2018detecting,lee2019structural,zevcevic2021interventional} to mention a select few), but also the establishment of new conferences such as CLeaR, suggests a growing subfield that sets a consensus on \emph{causal} AI/ML as promising paradigm for next-generation systems. Still, as the difficulty of the integration with otherwise prominent success stories of deep learning such as computer vision, with publications at CVPR and such (including \citep{sauer2021counterfactual,lv2022causality,liu2022structural} to mention a select few), becomes apparent, countering opinions start speaking out against causal AI/ML \citep{bishop2021artificial}. Nonetheless, online discussions on social media platforms like Twitter regarding the necessity of symbolic approaches, to which causality counts, for future learning systems \citep{marcus2021aifoundation} but also discussions during virtual gatherings as at the Causality Discussion Group \citep{brehmer2022cdg} all see researchers converging to a consensus on the promise given by causality for boosting AI/ML research towards its goal of intelligent systems. The aforementioned discussions polarize, however, since aspects such as how Pearlian causality might be understood in the realm of image data for instance is not yet clear as of writing. In this work, we take the arguably agreed upon perspective \emph{pro} causal AI/ML and we specifically try addressing intricacies and challenges that arise when viewing computer vision from a causal viewpoint.

\textbf{Structure of Paper Presentation.} In Sec.\ref{sec:2} we will present a way of viewing the PCH on image data, that is, for each of the levels of the hierarchy, what would corresponding images look like. In Sec.\ref{sec:3} we discuss four different intricacies to the perspective discussed prior that might come as surprising and that pose difficulties to finding models that can successfully generate a correct hierarchy on image data. In Sec.\ref{sec:4} we discuss two challenges (or tasks) of interest moving forward that our future models should be capable of completing successfully. In Sec.\ref{sec:5} we provide some last reflective thoughts on the presented introspection of the community's progress towards causal computer vision.

\section{Exemplifying the PCH on Images}\label{sec:2}
%[Start with observational and corgi as wrapfig. Then we add an intervention. However, this results in counterfactual which might be counterintuitive to some because we did an intervention - but it is actually correct, because everything else in image remained same. We also say that for large part this was the main confusion for many, since there are many works on interventions and counterfactuals in computer vision (cite). Also, how counterfactuals are part of the fundamental problem of causal inference and not usually observed. Now, we say, if other things can change that is a pure intervention. Finally, we conclude with the big overview which also shows it with a causal graph interpretation.]
Following the Pearlian notion to causality, the center model of study is the \emph{structural causal model} (SCM) which is defined as a 4-tuple $\mathcal{M}:=\inner{\mathbf{U},\mathbf{V},\mathcal{F},P(\mathbf{U})}$ where the so-called structural equations (which are deterministic functions) $v_i = f_i(\pa_i,u_i) \in \mathcal{F}$ assign values (denoted by lowercase letters) to the respective endogenous/system variables $V_i\in\mathbf{V}$ based on the values of their parents $\Pa_i\subseteq \mathbf{V}\setminus V_i$ and the values of some exogenous variables $\mathbf{U}_i\subseteq \mathbf{U}$ (sometimes also referred to as \emph{unmodelled} or \emph{nature} terms), and $P(\mathbf{U})$ denotes the probability function defined over $\mathbf{U}$. The SCM formalism comes with several interesting properties. They induce a causal graph $G$ and they induce what is known as the \emph{Pearl Causal Hierarchy} (PCH). See \citep{pearl2018book} for a first-time introduction to the topic. The hierarchy (sometimes also referred to as ladder) consists of three levels. The first, $\mathcal{L}_1$, is about observational/associational distributions over $\mathbf{V}$ with typical questions being ``What is?'', for example ``What does the symptoms tell us about the disease?''. While on $\mathcal{L}_1$ we are only concerned with a single observational distribution, on the second and the first causal level, $\mathcal{L}_2$, we have infinitely many interventional/hypothetical distributions with typical questions being ``What if?'', for example ``What if I take an aspirin, will my headache be cured?''. Finally and again infinitely many, counterfactual/retrospective distributions can be found on the third level, $\mathcal{L}_3$, with typical questions being ``Why?'', for example ``Was it the aspirin that cured my headache?''. A key result and sort of ``sanity check'' for research in causality was the establishment of the \emph{Causal Hierarchy Theorem} \citep{bareinboim20201on} which suggests (a) that any SCM will imply the PCH as just discussed with its $\mathcal{L}_1$ \emph{associational}, $\mathcal{L}_2$ \emph{interventional} and $\mathcal{L}_3$ \emph{counterfactual} levels, and more fundamentally (b) that causal quantities ($\mathcal{L}_i,i\in\{2,3\}$) are in fact richer in information than statistical quantities ($\mathcal{L}_1$), and that there exists a necessity of causal information (e.g.\ structural knowledge, essentially ``outside" model knowledge) for inference based on lower rungs e.g.\ $\mathcal{L}_1 \not\Rightarrow \mathcal{L}_2$ and therefore to reason about $\mathcal{L}_2$ or to \emph{identify} such causal quantities we need more than only observational data from $\mathcal{L}_1$. Interventions appear on both of the causal levels and are often times denoted with the $\doop$-notation. To conclude, consider the formal definition of valuations for the highest layer ($\mathcal{L}_3$) since it subsumes the other two layers as previously pointed out:
\begin{equation*} \label{eq:l12val}
    p(\mathbf{a}_{\mathbf{b}},\dots,\mathbf{c}_{\mathbf{d}}) = \textstyle\sum_{\mathcal{U}} p(\mathbf{u}) \quad \text{where} \quad \mathcal{U}=\{\mathbf{u}\mid \mathbf{A}_{\mathbf{b}}(\mathbf{u})=\mathbf{a},\dots, \mathbf{C}_{\mathbf{d}}(\mathbf{u})=\mathbf{c}\},
\end{equation*} for instantiations $\mathbf{a},\mathbf{b},\mathbf{c},\mathbf{d}$ of the node sets $\mathbf{A},\mathbf{B},\mathbf{C},\mathbf{D} \subseteq \mathbf{V}$ and they represent different ``worlds". The counterfactual $\mathbf{A}_{\mathbf{b}}(\mathbf{u})$ refers to the value $\mathbf{A}$ attains when $\mathbf{B}$ was deliberately \emph{set} to $\mathbf{b}$ in situation $\mathbf{u}$. E.g. for $\mathcal{L}_1$ we might only consider $\mathbf{A}=\mathbf{A}_{\emptyset}$ and whereas for $\mathcal{L}_2$ a single alternate world $\mathbf{A}_{\mathbf{b}}$. Note that the $\doop(\mathbf{x})$ notation and the counterfactual subscript notation $(\cdot)_{\mathbf{x}}$ do not generally coincide, they only coincide as $p(\mathbf{a}_{\mathbf{b}}\mid \mathbf{c}_{\mathbf{b}})=p(\mathbf{a}\mid \doop(\mathbf{b}),\mathbf{c})$ which is related to the discussion on whether the condition $\mathbf{c}$ is ``pre- or post-treatment'' (see \citet{pearl2009causality}).

\paragraph{$\mathcal{L}_1$ on Images.} In computer vision, we work with image data which is naturally represented as an ordered collection of pixel values commonly represented as matrices with multiple channels (e.g.\ RBG). Now, a common way of linking images with causality's SCM has been to have the SCM act as the image generating process. That is, the SCM will not represent pixels but rather ``high level concepts'' that nicely abstract the content and style of the images at hand. E.g.\ $\mathbf{V}$ could be specifying concepts like ``dog,'' which could be an indicator function suggesting that a ``dog'' should be placed within the image generated through that concrete specification. However, it is important to note that the SCM, as usual, is the sought after object of interest i.e., we do not know the model yet. In that sense, Pearlian causality comes in handy as a tool for abstraction and simply a language to formalizing the modelling assumptions. The first question we will answer now is, how do observations on image data, that is, instances of $\mathcal{L}_1$ for images look like? The image on the right illustrates an example.
\begin{wrapfigure}[9]{R}[30pt]{0.35\textwidth}
\vspace{-10pt}
\includegraphics[width=.25\textwidth]{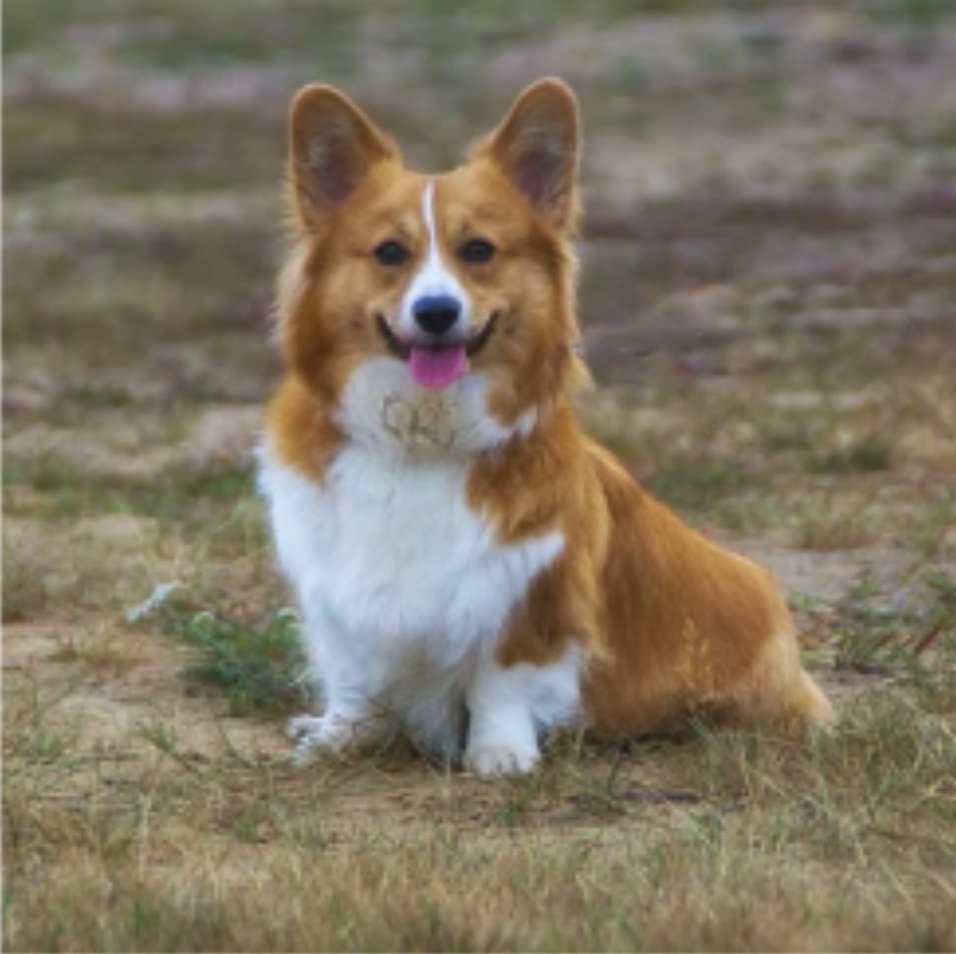}
\label{fig:l1}
\end{wrapfigure}

We can see a dog facing the image observer with its mouth open, tongue sticking out, ears upright, and sitting on a dry patch of lawn. The dog's fur is orange and white colored, the race of the dog is classified as corgi. This short description are all ``objective'' observations using human concepts about the appearance of the image that might or might not be captured formally in a corresponding SCM. Generally, we can state that $\mathcal{L}_1$ on images will simply correspond to any image collection that we consider to be our base set of images, however, with the important restriction that ``related'' images are excluded. As we will see shortly, a random set of images might very well contain such related images, which will naturally be a mix of $\mathcal{L}_1$ and $\mathcal{L}_2$ data.

\paragraph{$\mathcal{L}_3$ on Images.} What happens if you take the image of the corgi from before and \emph{change} aspects to it? Contrary to expectations, we end up on the final level of the PCH through an intervention on the \textit{same image}, $\mathcal{L}_3$. This might be counterintuitive and surprising since data augmentations are common practice in computer vision and usually associated with pure interventions ($\mathcal{L}_2$). See \citet{ilse2021selecting} for a discussion about data augmentations and Pearlian causality.
\begin{wrapfigure}[9]{R}[30pt]{0.35\textwidth}
\vspace{-10pt}
\includegraphics[width=.25\textwidth]{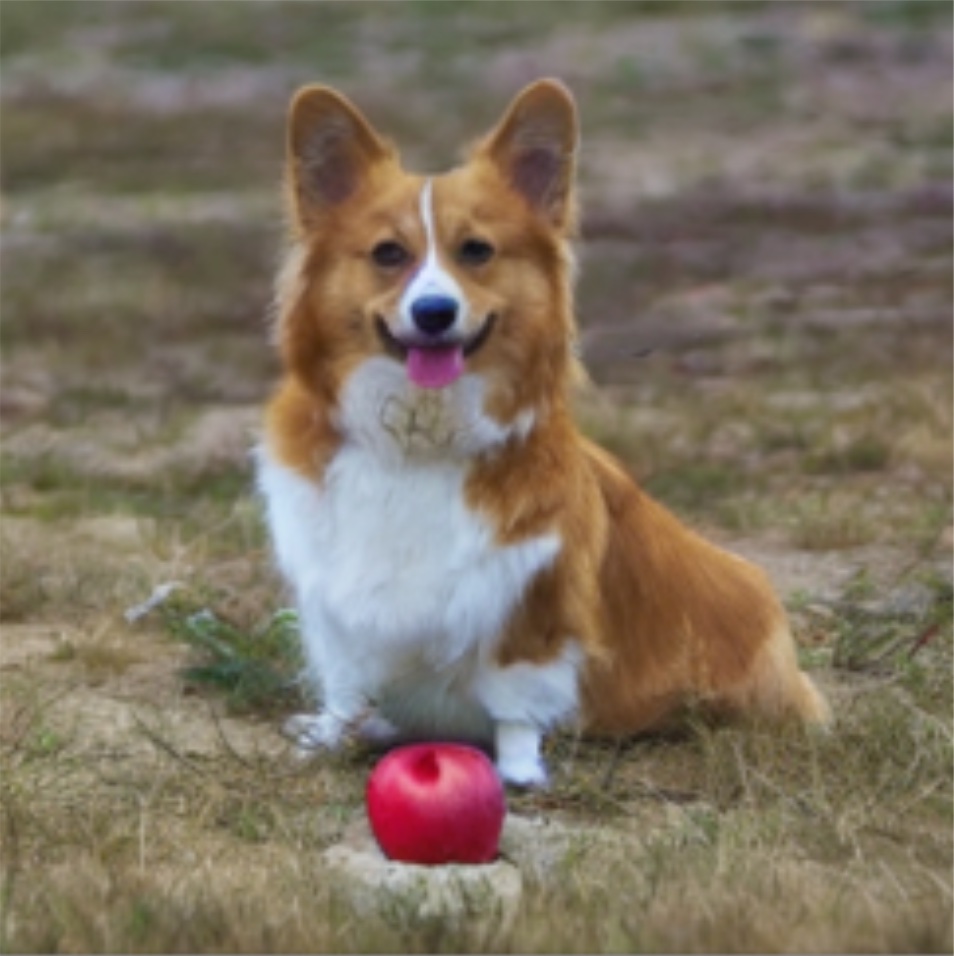}
\label{fig:l3}
\end{wrapfigure}

Using modern techniques from deep learning like stable diffusion \citep{rombach2022high} and/or a combination of inpainting and image editing we can generate a counterfactual image (countering the fact that we initially observed only the corgi) where there is suddenly a red apple placed right in front of the corgi on the lawn. In terms of an SCM that captures high level concepts on content, we can say that the exogenous terms $\mathbf{U}$ used for generating the actual images from the content indicators $\mathbf{V}$ are being kept fixed while the $\mathbf{V}$ has been modified in such way that the ``apple indicator'' is flipped on (that is, intervened upon).

\paragraph{$\mathcal{L}_2$ on Images.} To now obtain pure interventions, we need to relax the constraint that the new image would have to fix the $\mathbf{U}$ terms, that is, all the noise apart from the content indicators. Put simply, we don't have to observe the original, base image anymore. However, we still need to observe content for our given SCM in agreement with the base image, for example in this case a corgi should still be found in the intervened image.
\begin{wrapfigure}[9]{R}[30pt]{0.35\textwidth}
\vspace{-10pt}
\includegraphics[width=.25\textwidth]{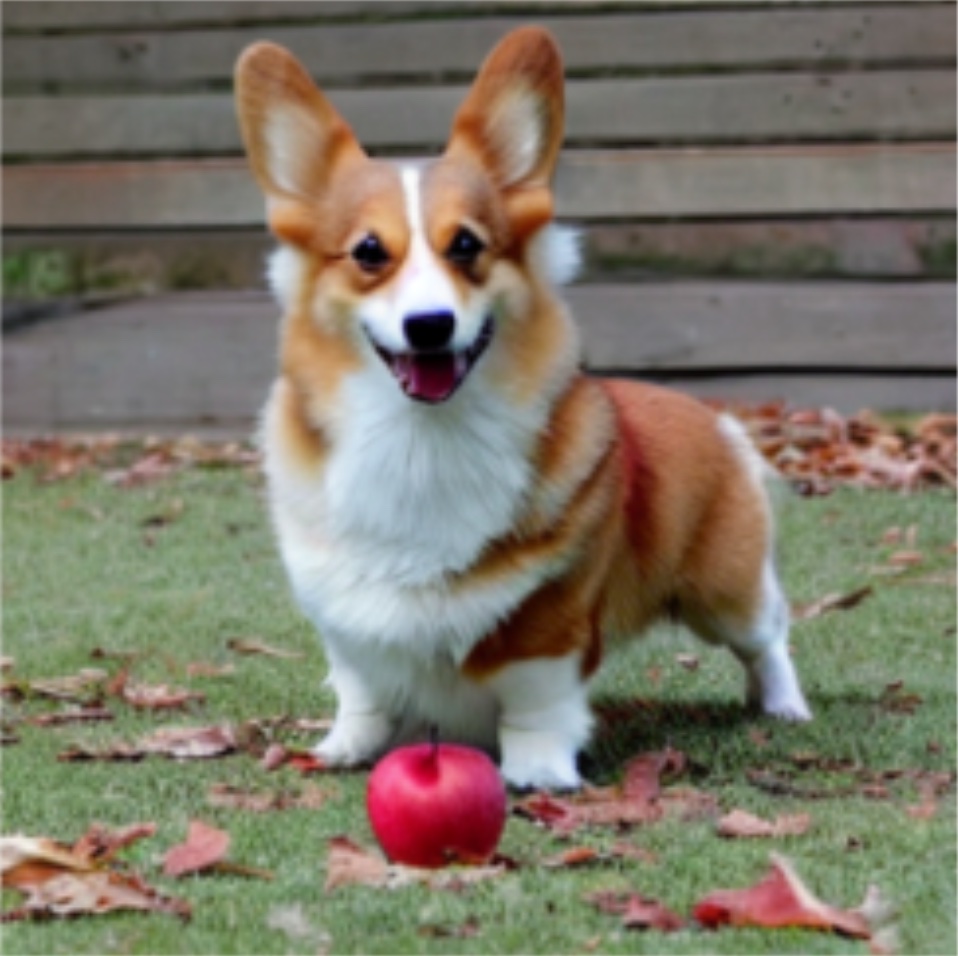}
\label{fig:l2}
\end{wrapfigure}

An example intervention is shown in the image on the right, which again presents a corgi and a red apple, but a different corgi on a different background thus not being constrained by the ``factual'' original image on all the unmodelled ``details'' of the image. We note two observations, (1) that the intervened image is difficult to synthesize through manual labour, that is, diffusion models open a new avenue of opportunity for creating true interventional image distributions, and (2) that the counterfactual from $\mathcal{L}_3$ can really be seen as very particular instance within the space of intervened images, the one that agrees with the base image on the unmodelled details $\mathbf{U}$.

\begin{figure}[t!]
\includegraphics[width=0.9\textwidth]{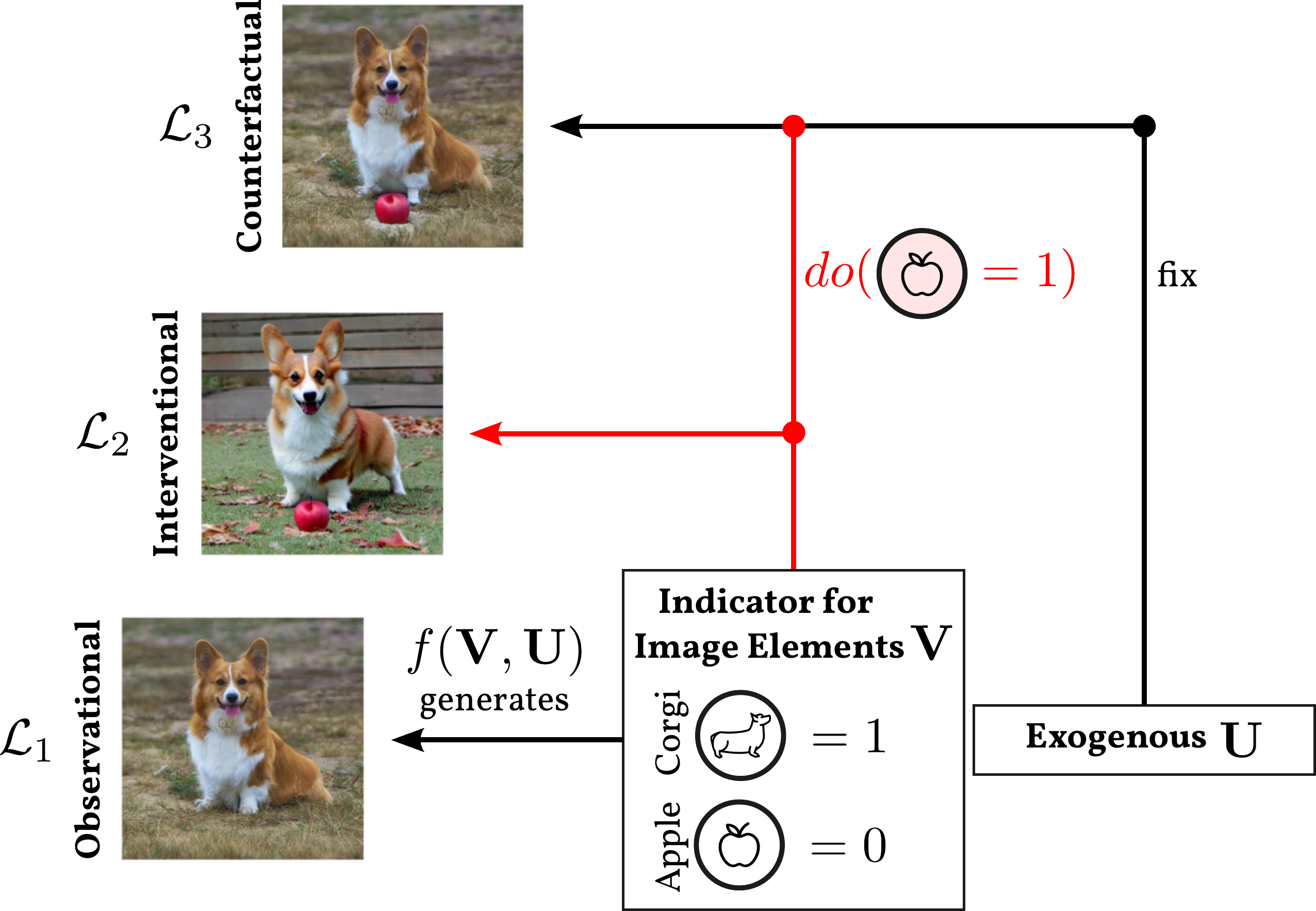}
\caption{\textbf{Pearl Causal Hierarchy on Image Data.} In the first level, $\mathcal{L}_1$, we are given some base image $\mathbf{X}_{\mathcal{L}_1}$ like the one shown that pictures a corgi facing the viewpoint. We can assume the SCM that generated this base image to capture high level binary indicators such as $V_1:=$``corgi in image'' or $V_2:=$``apple in image'' and for the base image, therefore, we observed $\mathbf{V}=(1,0)$. Naturally, unmodelled ``details'' $\mathbf{U}$ are concerned with specifications on what the corgi should look like, its pose, but also the background amongst others. Therefore, we have $\mathbf{X}_{\mathcal{L}_1}=f(\mathbf{V},\mathbf{U})$. To jump to the second level, $\mathcal{L}_2$, we can simply change $\mathbf{V}$ by intervening with $\doop(V_2=1)$ for instance, which leads to the appearance of a red apple in the intervened image $X_{\mathcal{L}_2}$. Note how the background and the corgi itself change through the intervention, this is because there is no restrection on the details $\mathbf{U}$ being placed by the intervention. That is, $\mathbf{U}$ is different for $X_{\mathcal{L}_1}$ and $X_{\mathcal{L}_2}$. Finally, on the third level, $\mathcal{L}_3$, we can re-introduce that restriction by fixing $\mathbf{U}$ and doing the intervention to return an image which looks the same as $X_{\mathcal{L}_1}$ up to the intervention of placing the apple. See Sec.\ref{sec:2} for a more detailed discussion. (Best viewed in color.)}
\label{fig:pchi}
\end{figure}

We summarize the insights collected in this section on the PCH for image data in a comprehensive schematic in Fig.\ref{fig:pchi} that highlights both the hierarchical nature and the requirements for commuting between the different levels.

\clearpage %%%
\section{Intricacies}\label{sec:3}
After our initial discussion on how one can interpret the PCH in terms of image data, exemplifying how images corresponding to each of the levels might look, we now move on to the finer details that come about naturally when being concerned with images, which complicate the overall problem of causal inference on images.

\subsection{Ambiguity of Interventions}\label{sec:31}
We have seen how interventions can be naturally formulated on the \emph{content} abstraction of images, that is, if there are high level concept indicators such as ``is there a dog in the image?'', then answering ``yes'' or ``no'' has predictable consequences in that a dog will or will not appear in the resulting image. However, what happens if we intervene on not what is being placed in the image but rather what is already within the image like, say, a bird sitting on a tree branch? If we do something like ``spread the wings of the bird'', then we surely underspecify what we mean by ``spreading wings.''
\begin{wrapfigure}[9]{R}[30pt]{0.6\textwidth}
\vspace{-10pt}
\includegraphics[width=.53\textwidth]{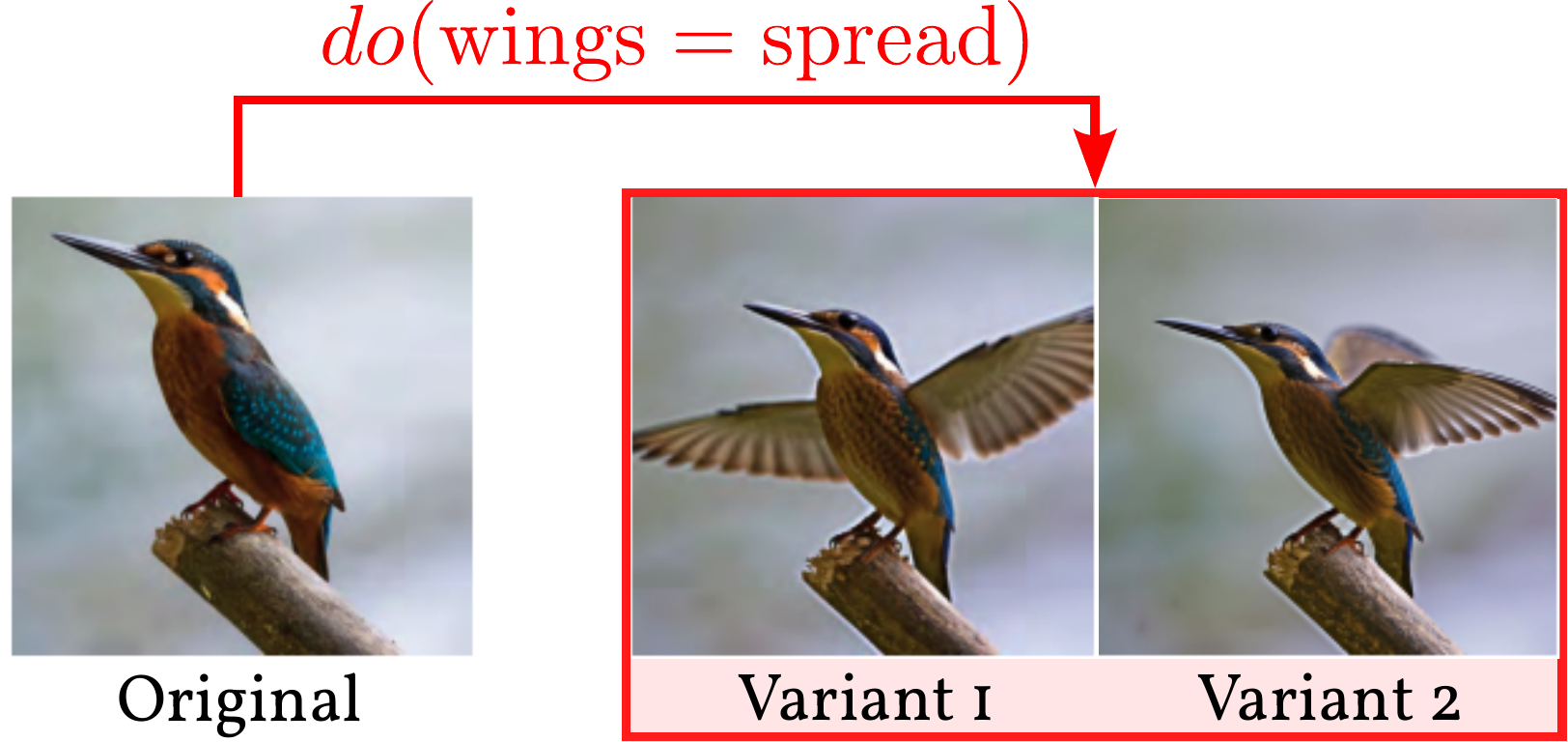}
%\caption{\textbf{Intricacy 1.} Description.}
\label{fig:i1}
\end{wrapfigure}

The figure on the right illustrates this idea that an intervention, especially in textual form as just given in the example, leaves open several aspects that somehow need to be decided for in the final image. In this example, the bird might spread the wings at an angle or to a certain range of motion as illustrated in variants 1 and 2, respectively. This observation becomes very apparent with cutting-edge methods in diffusion modeling currently under peer-review like Imagic \citep{kawar2022imagic} or Unitune \citep{valevski2022unitune}. 
%\begin{figure}
%\includegraphics[width=\textwidth]{Figures/intricacy-1.pdf}
%\caption{\textbf{Intricacy 1.} Description.}
%\label{fig:i1}
%\end{figure}

\subsection{Different Types of Interventions}\label{sec:32}
%[Here we really highlight the difference between "implicit" causality and independence. Where some interventions only change the image, whereas others change the causality that occurs in the image.]
In line with the discussion in the previous subsection, we could already see how ``not all interventions are created equal'' in that interventions can be of \emph{qualitatively} different nature.
\begin{wrapfigure}[15]{R}[50pt]{0.5\textwidth}
\vspace{-10pt}
\includegraphics[width=.34\textwidth]{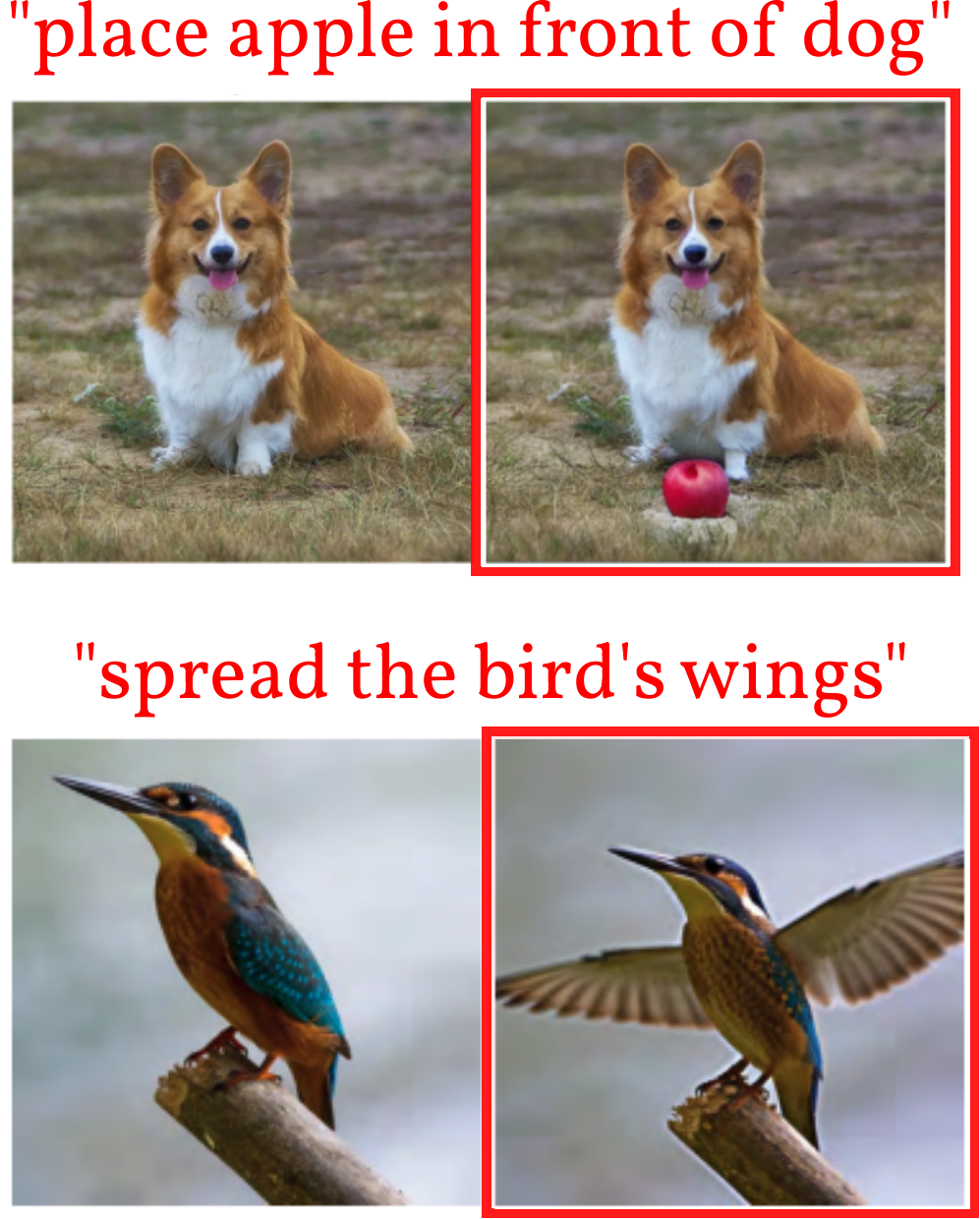}
%\caption{\textbf{Intricacy 2.} Description.}
\label{fig:i2}
\end{wrapfigure}

To be more precise, in the corgi example from Sec.\ref{sec:2} we simply placed an apple in front of the corgi, whereas in the example from Sec.\ref{sec:31} we let the bird in the image spread its wings. Consider the figure on the right that places the examples next to each other. The intervention in the top row places a new object within the image and naturally it also changes aspects of the previous image, for instance we cannot see the parts of the corgi or lawn that is being occluded by the newly placed apple. The intervention in the bottom row changes the state of the intervened variable, here the bird, in a way that is less local than previously in the object placement intervention. While it is difficult to capture the essence of what makes these interventions fundamentally different even in informal terms, we believe the main distinction to be that the object placement poses a pure intervention \emph{on the image}, whereas the wings spreading poses an intervention \emph{on the image elements}. This distinction we believe to be important because the latter contains a notion of \emph{implicit causation}, which can be seen as an indicator of the ``plausibility'' behind the image, for instance spreading the wings involves changing the bird's posture, the position of its limbs etc.\ To make this point more clear, imagine we placed a bone in the mouth of the corgi in the first example, this would constitute at first sight an independent object placement intervention, however, ideally, we expect the dog to hold it in its mouth by biting on it, which implies a causality similar to the second example with the bird. In any case, many of these aspects are up for debate, more importantly we wish to emphasize that the term ``intervention'' could already be ill-posed w.r.t.\ image data.
%\begin{figure}
%\includegraphics[width=\textwidth]{Figures/intricacy-2.pdf}
%\caption{\textbf{Intricacy 2.} Description.}
%\label{fig:i2}
%\end{figure}

\subsection{Inaccuracy of Counterfactuals}\label{sec:33}
Not only interventions, but also counterfactuals introduce issues that complicate causal inference on images. A counterfactual, as defined by Pearl, subsumes a three step procedure that involves updating your beliefs on the exogenous $\mathbf{U}$ but also performing interventions. We have seen ambiguities caused by interventions but they also come with the exogenous details unspecified by the given SCM.
\begin{wrapfigure}[16]{R}[50pt]{0.6\textwidth}
\includegraphics[width=.48\textwidth]{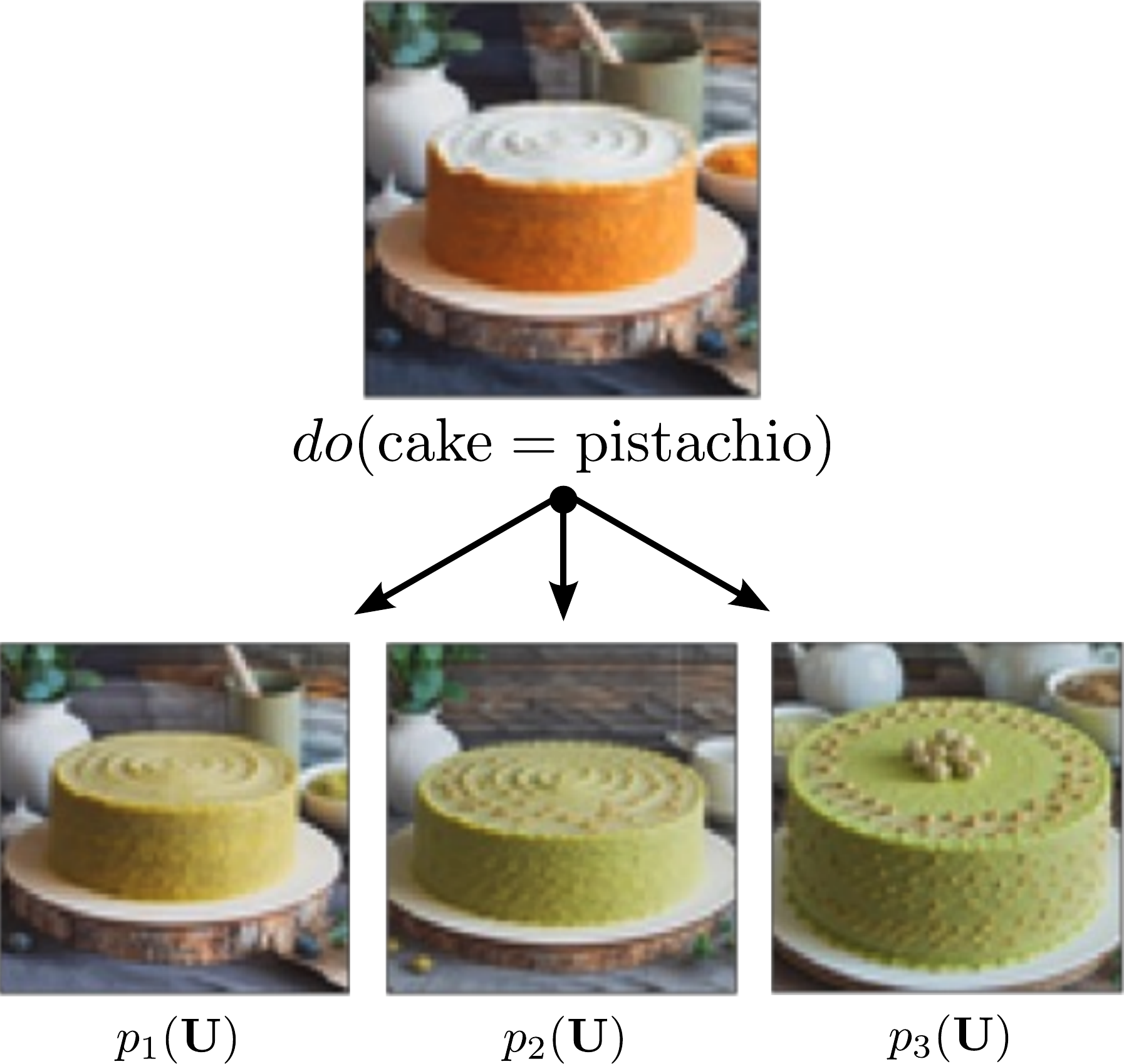}
%\caption{\textbf{Intricacy 3.} Description.}
\label{fig:i3}
\end{wrapfigure}

Consider the example given in the figure on the right, where the base image of a cake is being intervened upon such that the new image contains an image of a pistachio cake. Ideally, the counterfactual should be the same as the base image up to the intervention. Yet, as we realize from our previous discussions, what constitutes ``same'' is not clear, that is, where do we \emph{exactly} find all the aspects of the exogenous terms within the image? The example highlights three different distributions over $\mathbf{U}$, representing different beliefs (or conceptions), of the exogenous terms and what ``same'' \emph{should} mean. We can clearly see that there is no ideal answer i.e., out of all three distribution $p_1(\mathbf{U})$ keeps the surrounding most constant (e.g.\ by looking at the wooden serving plate and the similar viewpoint) but is arguably also the ``least pistachio'' like out of the three. Same arguments can be made for the other two, e.g.\ $p_3(\mathbf{U})$ even contains pieces of pistachio on top of the cake but the background changed significantly introducing new dishware and also a very different viewing angle additionally, or for instance $p_2(\mathbf{U})$ seems to compromise between the two previous but is significantly wider than the cake from the base image. The highlighted issue of inaccuracy in counterfactuals becomes especially prevalent in methods such as Imagic \citep{kawar2022imagic} that make use of thresholding to simply settle for a decided notion of proximity which likely is or even needs to be guided by human supervision. To move forward, we believe that any of three beliefs about what constitutes $\mathbf{U}$ would have been sufficient for most relevant tasks of interest, nonetheless, we intend on raising awareness to this intricacy which might otherwise would have fallen under the radar, since the ambiguity otherwise naturally found in images also naturally translates to a proximity issue for counterfactual images.
%\begin{figure}
%\includegraphics[width=\textwidth]{Figures/intricacy-3.pdf}
%\caption{\textbf{Intricacy 3.} Description.}
%\label{fig:i3}
%\end{figure}

\subsection{Inpainting versus Fine-Tuning}\label{sec:34}
In this last subsection on intricacies, we intend on shortly discussing two existing methods with which interventional and counterfactual images can be generated right now: inpainting (IP) and fine-tuning of existing diffusion models (FT). In IP we generally have two options we can either mask out certain image segments like for instance the mouth of the corgi from the first example and then command a diffusion model to fill in the newly created gap by ``placing a bone in the corgi's mouth'', or we can directly use photo editing software to modify the original image. In FT on the other hand, we might deploy an approach like Imagic \citep{kawar2022imagic} or Unitune \citep{valevski2022unitune} where the authors fine-tune existing diffusion models to regularize the embedding space in such way that for any given specific intervention the synthesized images adhere more strongly to what is being queried by the intervention. For example in a method such as SDEdit \citep{meng2021sdedit} or Text2Live \citep{bar2022text2live} the synthesized images might ignore the intervention altogether or even worse, corrupt the base image. While both IP and FT are techniques readily available for reproduction (even when models like for Imagic and Unitune are unfortunately being kept disclosed under intellectual property rights) and could thus be deployed to generate a \emph{truly} causal data set consisting of images from all three levels of the PCH, still, it becomes apparent that practically this is infeasible since both IP and FT are costly methods for generating causal data, especially at the scale of modern techniques of deep learning. For IP, either manual labour or the restriction to certain types of interventions (like object placement) would be necessary to synthesize interventional and counterfactual images, while for FT, fine-tuning for each generation would be necessary, thereby rendering both approaches practically irrelevant for synthesizing causal image data sets.

\section{Challenges: Tasks our Future Models Need to Solve}\label{sec:4}
In the previous section we discussed several intricacies that also pose challenges to causal inference with images, but now we turn our attention to actual challenges in the sense of \emph{tasks that we wish our models to be capable of solving}. We present two different challenges (or tasks) that arise naturally as a consequence of our interpretation of the PCH for image data.

\paragraph{Previous Attempts at Causal Image Data Sets.} There have been several strides in coming up with a causal or ``kind-of'' causal data set. For instance, \citet{gondal2019transfer} looked at disentanglement and factors of variation and created both synthetic and real-world images of objects in different poses, colors, etc. In \citep{singla2021causal}, the authors used manual labour using a mechanical turk study to discover spurious features in images, while \citet{lynch2022evaluating} recently employed diffusion models and base assumptions on the underlying SCM to synthesize data sets that balance out each of the existing high level concepts (e.g.\ if there was never a cow on the beach in the original data set, then now there was). All of these aforementioned approaches mitigate some of the core issues with causal data sets, especially for images. However, none of them has been able to fully subsume the PCH, and unfortunately, due to the difficulties with existing techniques as highlighted in Sec.\ref{sec:34}, a truly causal image data set seems to still be out of reach as not even brute force techniques could force enough data samples for modern ML techniques lingering for ever-increasing data amounts. Nonetheless, we believe that the data set would constitute only part of the overall big picture. What do we do when given such a truly causal image data set? Next we try providing an answer to this question by highlighting challenges/tasks for our future models.

\subsection{Imagining a Counterfactual Image}\label{sec:41}
\begin{figure}[t!]
\centering
\includegraphics[width=.85\textwidth]{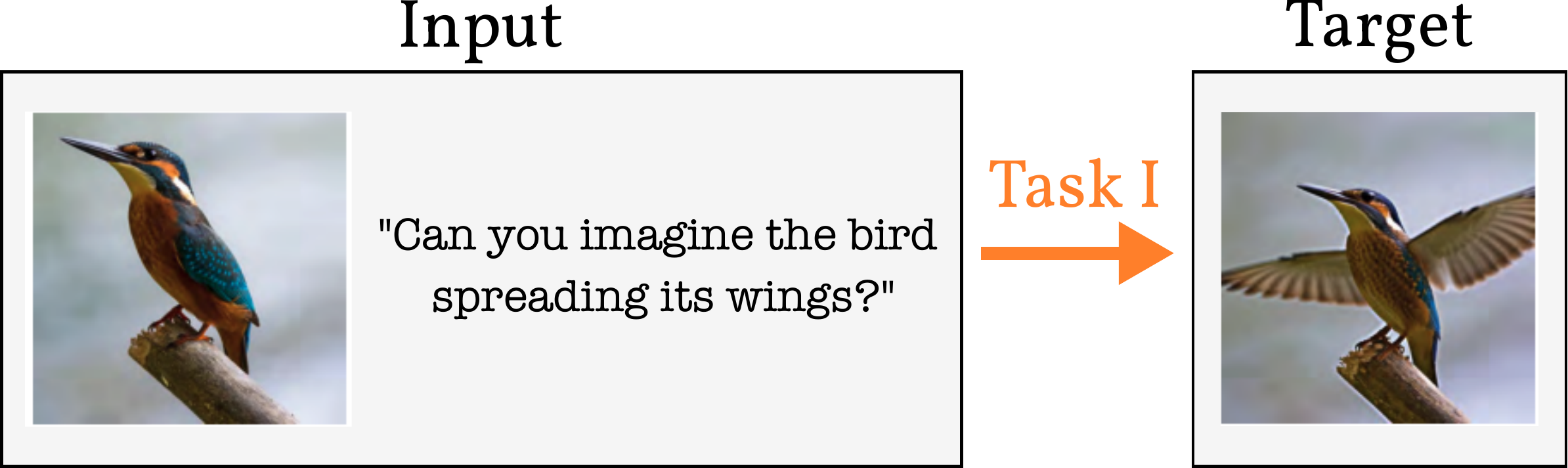}
\caption{\textbf{Task I, ``Imagining Counterfactuals.''} Given an input tuple containing an image and an intervention, \emph{synthesize} the corresponding counterfactual image. (Best viewed in color.)}
\label{fig:c1}
\end{figure}

The task is being illustrated schematically in Fig.\ref{fig:c1} with the bird example from previous discussions. Counterfactuals pose, in Pearl's term, the center piece to his theory of causation and are correspondingly placed on the highest level (or rung) of the PCH. Opposed to pure interventions which talk about hypothetical situations, counterfactuals are all about \emph{retrospection}. That is, through knowing about the exogenous terms $\mathbf{U}$, which in the standard case imply a certain $\mathbf{V}=\mathbf{v}$ since fixing $\mathbf{U}=\mathbf{u}$ results in deterministic functions $\mathbf{v}=f(\mathbf{pa},\mathbf{u})$, we move ``back in time'' and consider an alternate \emph{world} configuration. This observation also gives rise to Pearl's opinion that counterfactuals corresponding to ``understanding'' and the ``highest mode of human cognition'' (\citet{pearl2018book} further illustrate $\mathcal{L}_3$ with a cartoon sketch of Einstein reflecting on complex inventions such as rockets or laptops). As of now, diffusion models are capable of creating counterfactuals (as clearly the given examples in this paper illustrate, since they were indeed generated with the help of diffusion models), however, they require fine-tuning or inpainting, which effectively changes the original model (which amounts to defeating the purpose of the challenge). Therefore, since counterfactual reasoning is desirable, and current models are not able to generate counterfactuals in a \textit{targeted manner}, we propose in Task I to ``imagine counterfactuals.'' That is, given an input tuple containing an image and an intervention, synthesize the corresponding counterfactual image. In order to succeed in this task the queried system needs to recognize the point of intervention (here: bird, and that specific bird if there are multiple) and also ``know'' what the consequences of the given intervention are (here: limbs moving, posture changing etc.\ due to the wing spread).

\subsection{Identifying the Interventions within Images}\label{sec:42}
\begin{figure}[t!]
\centering
\includegraphics[width=.85\textwidth]{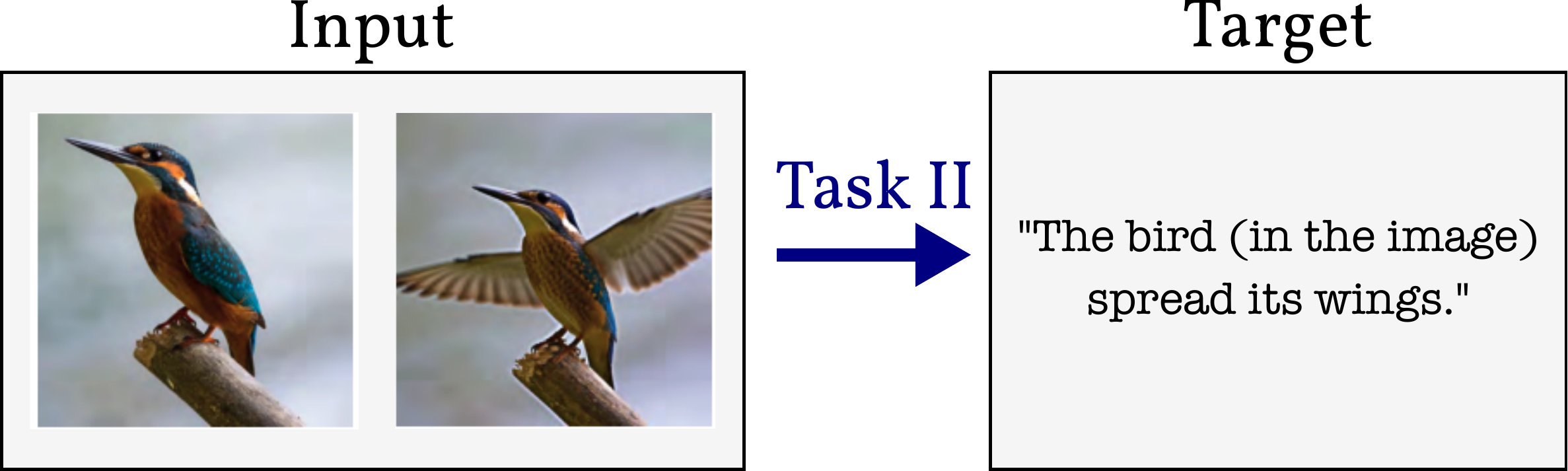}
\caption{\textbf{Task II, ``Identifying Interventions.''} Given an input tuple containing a base image and a corresponding image in which an intervention occured, \emph{extract} the corresponding intervention. (Best viewed in color.)}
\vspace{-.5cm}
\label{fig:c2}
\end{figure}

The intricate relation between interventions and counterfactuals becomes more apparent in the second task that we propose, which is illustrated in Fig.\ref{fig:c2}, again with the bird example. Forming a counterfactual involves performing an intervention in a fixed-$\mathbf{U}$ world. The task is specified as follows: given an input tuple containing a base image and a corresponding image in which an intervention occured, extract the corresponding intervention. The intervention is implicitly hidden within the image and the queried system needs to be able to identify the change between the images and explicate the \emph{reason} for the change (here: the bird spreading its wing). While in Task I the queried system is concerned with the same hypothesis space (that of images) for providing the answer, in Task II the multi-modality switches the answering space to be in natural language. Again, like with Task I, existing state-of-the-art methods for image captioning like X-LAN \citep{pan2020x} are capable of capturing precisely what is ``going on'' in a given image, and thus a naïve solution to Task II would be to simply look at some sensible notion of difference between the two independently captioned images. However, we'd again be defeating the purpose since we want to have an \emph{inherently single} model being capable of solving this task by processing the input pair through some internal representation in such way that the intervention becomes apparent.

\paragraph{Solving Tasks I and II.} If we had a model that can solve Task I and II, then we'd have a model that can (a) understand interventions and (b) use them to reason counterfactually, and all that within images. Nonetheless, a very important aspect to this that fell short in the previous discussions is the aspect of time or dynamics unrolling. We humans know how it looks like when a bird is spreading its wings, and so we are able to solve both Tasks I and II gracefully although being confronted with \emph{still} images (with the asteriks on Task I because a human, without aid, will not provide the pixel values for the counterfactual image). We learn to explore the world interactively in a Markov Decision Process type situation (see for instance the \citet{bareinboim2020talk} ICML Tutorial on Causal Reinforcement Learning). Therefore, we need to be aware of the assumptions and restrictions we place on the learning problem for our models when facing Tasks I and II.

\section{Concluding Remarks}\label{sec:5}
The belief in the utility of causality for AI/ML permeates the ongoing narrative, uniting the community in the hope that causal models hold a significant part of the future of intelligent systems. However, there is still a bitter taste left for many, since they do agree on the fact that being able to answer causal queries of the interventional and counterfactual nature is desirable, but worry about the promise never being delivered. This feeling is only being further corroborated by a lack of existing data sets, especially in domains that have historically been the success stories of AI/ML like computer vision. But also through a lack of clarity in how the intersection of causality and computer vision is to be understood. In this work we highlighted how we can interpret and project the Pearl Causal Hierarchy on image data, and subtle intricacies that come along with it. Furthermore, we discussed challenges that involve tasks our future image understanding systems should certainly be capable of handling. We hope that this work can inspire researchers by painting a vivid picture of key ideas and emphasizing important avenues for future work.

\clearpage
\subsubsection*{Acknowledgments}
The authors acknowledge the support of the German Science Foundation (DFG) project “Causality, Argumentation, and Machine Learning” (CAML2, KE 1686/3-2) of the SPP 1999 “Robust Argumentation Machines” (RATIO). This work was supported by the ICT-48 Network of AI Research Excellence Center “TAILOR” (EU Horizon 2020, GA No 952215), the Nexplore Collaboration Lab “AI in Construction” (AICO) and by the Federal Ministry of Education and Research (BMBF; project “PlexPlain”, FKZ 01IS19081). It benefited from the Hessian research priority programme LOEWE within the project WhiteBox \& the HMWK cluster project “The Third Wave of AI” (3AI).
\bibliography{references}

\begin{thebibliography}{26}
\providecommand{\natexlab}[1]{#1}
\providecommand{\url}[1]{\texttt{#1}}
\expandafter\ifx\csname urlstyle\endcsname\relax
  \providecommand{\doi}[1]{doi: #1}\else
  \providecommand{\doi}{doi: \begingroup \urlstyle{rm}\Url}\fi

\bibitem[Bar-Tal et~al.(2022)Bar-Tal, Ofri-Amar, Fridman, Kasten, and
  Dekel]{bar2022text2live}
Omer Bar-Tal, Dolev Ofri-Amar, Rafail Fridman, Yoni Kasten, and Tali Dekel.
\newblock Text2live: Text-driven layered image and video editing.
\newblock \emph{arXiv preprint arXiv:2204.02491}, 2022.

\bibitem[Bareinboim(2020)]{bareinboim2020talk}
Elias Bareinboim.
\newblock Tutorial on ``towards causal reinforcement learning'', 2020.
\newblock URL \url{https://www.youtube.com/watch?v=QRTgLWfFBMM}.

\bibitem[Bareinboim et~al.(2020)Bareinboim, Correa, Ibeling, and
  Icard]{bareinboim20201on}
Elias Bareinboim, Juan~D Correa, Duligur Ibeling, and Thomas Icard.
\newblock 1on pearl’s hierarchy and.
\newblock Technical report, Technical Report, 2020.

\bibitem[Bishop(2021)]{bishop2021artificial}
J~Mark Bishop.
\newblock Artificial intelligence is stupid and causal reasoning will not fix
  it.
\newblock \emph{Frontiers in Psychology}, 11:\penalty0 2603, 2021.

\bibitem[Brehmer(2022)]{brehmer2022cdg}
Johann Brehmer.
\newblock Causality discussion group session on 12.10.2022, 2022.
\newblock URL \url{https://www.youtube.com/watch?v=u38E6OXzuwY}.

\bibitem[Geffner et~al.(2022)Geffner, Dechter, and
  Halpern]{geffner2022probabilistic}
Hector Geffner, Rina Dechter, and Joseph~Y Halpern.
\newblock Probabilistic and causal inference: The works of judea pearl, 2022.

\bibitem[Gondal et~al.(2019)Gondal, Wuthrich, Miladinovic, Locatello, Breidt,
  Volchkov, Akpo, Bachem, Sch{\"o}lkopf, and Bauer]{gondal2019transfer}
Muhammad~Waleed Gondal, Manuel Wuthrich, Djordje Miladinovic, Francesco
  Locatello, Martin Breidt, Valentin Volchkov, Joel Akpo, Olivier Bachem,
  Bernhard Sch{\"o}lkopf, and Stefan Bauer.
\newblock On the transfer of inductive bias from simulation to the real world:
  a new disentanglement dataset.
\newblock \emph{Advances in Neural Information Processing Systems}, 32, 2019.

\bibitem[Ilse et~al.(2021)Ilse, Tomczak, and Forr{\'e}]{ilse2021selecting}
Maximilian Ilse, Jakub~M Tomczak, and Patrick Forr{\'e}.
\newblock Selecting data augmentation for simulating interventions.
\newblock In \emph{International Conference on Machine Learning}, pages
  4555--4562. PMLR, 2021.

\bibitem[Janzing and Sch{\"o}lkopf(2018)]{janzing2018detecting}
Dominik Janzing and Bernhard Sch{\"o}lkopf.
\newblock Detecting non-causal artifacts in multivariate linear regression
  models.
\newblock In \emph{International Conference on Machine Learning}, pages
  2245--2253. PMLR, 2018.

\bibitem[Kawar et~al.(2022)Kawar, Zada, Lang, Tov, Chang, Dekel, Mosseri, and
  Irani]{kawar2022imagic}
Bahjat Kawar, Shiran Zada, Oran Lang, Omer Tov, Huiwen Chang, Tali Dekel, Inbar
  Mosseri, and Michal Irani.
\newblock Imagic: Text-based real image editing with diffusion models.
\newblock \emph{arXiv preprint arXiv:2210.09276}, 2022.

\bibitem[Lee and Bareinboim(2019)]{lee2019structural}
Sanghack Lee and Elias Bareinboim.
\newblock Structural causal bandits with non-manipulable variables.
\newblock In \emph{Proceedings of the AAAI Conference on Artificial
  Intelligence}, volume~33, pages 4164--4172, 2019.

\bibitem[Liu et~al.(2022)Liu, Liu, Paull, Weller, and
  Sch{\"o}lkopf]{liu2022structural}
Weiyang Liu, Zhen Liu, Liam Paull, Adrian Weller, and Bernhard Sch{\"o}lkopf.
\newblock Structural causal 3d reconstruction.
\newblock In \emph{European Conference on Computer Vision}, pages 140--159.
  Springer, 2022.

\bibitem[Lv et~al.(2022)Lv, Liang, Li, Zang, Liu, Wang, and
  Liu]{lv2022causality}
Fangrui Lv, Jian Liang, Shuang Li, Bin Zang, Chi~Harold Liu, Ziteng Wang, and
  Di~Liu.
\newblock Causality inspired representation learning for domain generalization.
\newblock In \emph{Proceedings of the IEEE/CVF Conference on Computer Vision
  and Pattern Recognition}, pages 8046--8056, 2022.

\bibitem[Lynch et~al.(2022)Lynch, Kaddour, and Silva]{lynch2022evaluating}
Aengus Lynch, Jean Kaddour, and Ricardo Silva.
\newblock Evaluating the impact of geometric and statistical skews on
  out-of-distribution generalization performance.
\newblock In \emph{NeurIPS 2022 Workshop on Distribution Shifts: Connecting
  Methods and Applications}, 2022.

\bibitem[Marcus and Davis(2021)]{marcus2021aifoundation}
Gary Marcus and Ernest Davis.
\newblock Has ai found a new foundation?
\newblock \emph{The Gradient}, 2021.

\bibitem[Meng et~al.(2021)Meng, He, Song, Song, Wu, Zhu, and
  Ermon]{meng2021sdedit}
Chenlin Meng, Yutong He, Yang Song, Jiaming Song, Jiajun Wu, Jun-Yan Zhu, and
  Stefano Ermon.
\newblock Sdedit: Guided image synthesis and editing with stochastic
  differential equations.
\newblock In \emph{International Conference on Learning Representations}, 2021.

\bibitem[Pan et~al.(2020)Pan, Yao, Li, and Mei]{pan2020x}
Yingwei Pan, Ting Yao, Yehao Li, and Tao Mei.
\newblock X-linear attention networks for image captioning.
\newblock In \emph{Proceedings of the IEEE/CVF conference on computer vision
  and pattern recognition}, pages 10971--10980, 2020.

\bibitem[Pearl(2009)]{pearl2009causality}
Judea Pearl.
\newblock \emph{Causality}.
\newblock Cambridge university press, 2009.

\bibitem[Pearl and Mackenzie(2018)]{pearl2018book}
Judea Pearl and Dana Mackenzie.
\newblock \emph{The book of why: the new science of cause and effect}.
\newblock Basic books, 2018.

\bibitem[Peters et~al.(2017)Peters, Janzing, and
  Sch{\"o}lkopf]{peters2017elements}
Jonas Peters, Dominik Janzing, and Bernhard Sch{\"o}lkopf.
\newblock \emph{Elements of causal inference: foundations and learning
  algorithms}.
\newblock The MIT Press, 2017.

\bibitem[Rombach et~al.(2022)Rombach, Blattmann, Lorenz, Esser, and
  Ommer]{rombach2022high}
Robin Rombach, Andreas Blattmann, Dominik Lorenz, Patrick Esser, and Bj{\"o}rn
  Ommer.
\newblock High-resolution image synthesis with latent diffusion models.
\newblock In \emph{Proceedings of the IEEE/CVF Conference on Computer Vision
  and Pattern Recognition}, pages 10684--10695, 2022.

\bibitem[Sauer and Geiger(2021)]{sauer2021counterfactual}
Axel Sauer and Andreas Geiger.
\newblock Counterfactual generative networks.
\newblock \emph{arXiv preprint arXiv:2101.06046}, 2021.

\bibitem[Sch{\"o}lkopf(2022)]{scholkopf2022causality}
Bernhard Sch{\"o}lkopf.
\newblock Causality for machine learning.
\newblock In \emph{Probabilistic and Causal Inference: The Works of Judea
  Pearl}, pages 765--804. 2022.

\bibitem[Singla and Feizi(2021)]{singla2021causal}
Sahil Singla and Soheil Feizi.
\newblock Causal imagenet: How to discover spurious features in deep learning?
\newblock \emph{arXiv preprint arXiv:2110.04301}, 2021.

\bibitem[Valevski et~al.(2022)Valevski, Kalman, Matias, and
  Leviathan]{valevski2022unitune}
Dani Valevski, Matan Kalman, Yossi Matias, and Yaniv Leviathan.
\newblock Unitune: Text-driven image editing by fine tuning an image generation
  model on a single image.
\newblock \emph{arXiv preprint arXiv:2210.09477}, 2022.

\bibitem[Ze{\v{c}}evi{\'c} et~al.(2021)Ze{\v{c}}evi{\'c}, Dhami, Karanam,
  Natarajan, and Kersting]{zevcevic2021interventional}
Matej Ze{\v{c}}evi{\'c}, Devendra Dhami, Athresh Karanam, Sriraam Natarajan,
  and Kristian Kersting.
\newblock Interventional sum-product networks: Causal inference with tractable
  probabilistic models.
\newblock \emph{Advances in Neural Information Processing Systems}, 34, 2021.

\end{thebibliography}

%\appendix

\end{document}